\documentclass{article}
\usepackage{ijcai17}

\usepackage{times}

\usepackage{amsmath}

\usepackage{courier}
\usepackage{ensuretext}
\usepackage{mathnotation}
\usepackage{amsfonts}
\usepackage{algorithm}
\usepackage{verbatim}
\usepackage[noend]{algpseudocode}
\algrenewcommand\algorithmicrequire{\textbf{Input:}}
\algrenewcommand\algorithmicensure{\textbf{Output:}}
\usepackage{color}	
\definecolor{light-gray1}{gray}{0.95}
\usepackage[table]{xcolor}
\usepackage{amsthm}
\usepackage{url}
\usepackage{graphicx}
\usepackage{multirow}
\usepackage{balance}
\usepackage{flushend}
\usepackage{xspace}

\usepackage[inline]{enumitem}

\usepackage{threeparttable}

\newcommand{\md}{\mathcal{D}}

\newcommand{\Tp}{\mathit{P}\xspace}
\newcommand{\Tn}{\mathit{N}\xspace}
\newcommand{\tp}{\mathit{p}}
\newcommand{\tn}{\mathit{n}}

\newcommand{\LD}{{\bf{D}}}
\newcommand{\allD}{{\mathfrak{D}}}
\newcommand{\dx}[1]{{\bf D}_{#1}^+}
\newcommand{\dnx}[1]{{\bf D}_{#1}^{-}}
\newcommand{\dz}[1]{{\bf D}_{#1}^0}
\newcommand{\mQ}{{\bf{Q}}}
\newcommand{\mD}{{\bf{D}}}
\newcommand{\Pt}{\mathfrak{P}}
\newcommand{\Disc}{\mathsf{Disc}}

\newcommand{\sd}{\textsc{sd}\xspace}
\newcommand{\comps}{\textsc{comps}\xspace}
\newcommand{\obs}{\textsc{obs}\xspace}
\newcommand{\ab}{\textsc{ab}}

\newcommand{\sdaa}[1]{\textsc{sd}^*[#1]}
\newcommand{\beh}[1]{\textsc{beh}[#1]}

\newcommand{\dg}{\ensuremath{\Delta}\xspace}

\newtheorem{definition}{Definition}
\newtheorem{theorem}{Theorem}
\newtheorem{proposition}{Proposition}

\newtheorem{property}{Property}
\newtheorem{problem}{Problem}
\newtheorem{conjecture}{Conjecture}

\title{Inexpensive Cost-Optimized Measurement Proposal \\ for Sequential Model-Based Diagnosis}
\author%
{%
	Patrick Rodler$^1$ \and Wolfgang Schmid$^1$ \and Konstantin Schekotihin$^1$\\
	$^1$Alpen-Adria Universit\"at, Klagenfurt, Austria\\
	e-mail: firstname.lastname@aau.at\\
}

\begin{document}
	%
	
	
	\maketitle
	\begin{abstract}
In this work we present strategies for (optimal) measurement selection in model-based sequential diagnosis. 
In particular, assuming a set of leading diagnoses being given, we show how queries (sets of measurements) can be computed and optimized along two dimensions: expected number of queries and cost per query. 
By means of a suitable decoupling of two optimizations and a clever search space reduction the computations are done without any inference engine calls. 
For the full search space, we give a method requiring only a polynomial number of inferences and guaranteeing query properties existing methods cannot provide. 
Evaluation results using real-world problems indicate that the new method computes (virtually) optimal queries instantly independently of the size and complexity of the considered diagnosis problems.
\end{abstract}
	
	
\section{Introduction} \label{sec:intro}

Model-based diagnosis (MBD) is a widely applied approach to finding explanations, called \emph{diagnoses}, for unexpected behavior of observed systems including hardware, software, knowledge bases, discrete event systems, feature models, user interfaces, etc.\ \cite{Reiter87,dressler1996consistency,DBLP:conf/aadebug/MateisSWW00,pencole2005formal,Kalyanpur.Just.ISWC07,DBLP:journals/apin/FelfernigFISTJ09,DBLP:journals/jss/WhiteBSTDC10}. 
In case the provided observations are insufficient for successful fault localization, \emph{sequential diagnosis (SQD)} methods 
collect additional information by generating a sequence of \emph{queries} \cite{dekleer1987,pietersma2005model,DBLP:journals/jair/FeldmanPG10a,Siddiqi2011,Shchekotykhin2012}.\footnote{Following the arguments of \cite{pietersma2005model} we do not consider non-MBD sequential methods~\cite{pattipati1990,DBLP:journals/tsmc/ShakeriRPP00,DBLP:journals/mam/ZuzekBN00,DBLP:conf/ijcai/BrodieRMO03}.}
As query answering is often costly, the goal of SQD is to \emph{minimize the diagnostic cost}, like time or manpower, required to achieve a diagnostic goal, e.g.\ a highly probable diagnosis.
%
To this end, the cited SQD works minimize the \emph{number of queries} by a one-step lookahead measure $m$, e.g.\ entropy \cite{dekleer1987}, but do not optimize the \emph{query cost}, such as the time required to perform measurements~\cite{heckerman1995decision}.
  
%

\noindent\textbf{Contributions.} 
We present a novel query optimization method that is generally applicable to any MBD problem in the sense of \cite{dekleer1987,Reiter87} and \\
	\noindent\textbf{(1)} defines a query as a set of first-order sentences and thus generalizes the \emph{measurement} notion of \cite{dekleer1987,Reiter87}, \\
	\noindent\textbf{(2)} given a set of \emph{leading} 
	\emph{diagnoses} \cite{DBLP:conf/ijcai/KleerW89}, allows the two-dimensional optimization of the next query in terms of the \emph{expected number of subsequent queries} (measure $m$) and \emph{query cost} (measure $c$), \\
	%
	\noindent\textbf{(3)} for an aptly refined (yet exponential) query search space, finds -- \emph{without any reasoner calls} -- the \emph{globally} optimal query wrt.\ measure $c$ that \emph{globally} optimizes measure $m$, \\
	%
		%
	\noindent\textbf{(4)} for the full query search space, finds -- with a polynomial number of reasoner calls -- the (under reasonable assumptions) globally optimal query wrt.\ $m$ that includes, if possible, only ``cost-preferred'' sentences, such as those answerable automatically using built-in sensors, \\
	\noindent\textbf{(5)} guarantees the proposal of queries that discriminate between \emph{all} leading diagnoses and that \emph{unambiguously identify the actual diagnosis}. 

The efficiency of our approach is possible by the recognition that the optimizations of $m$ and $c$ can be decoupled and by using logical monotonicity as well as the inherent (already inferred) information in the ($\subseteq$-minimal) leading diagnoses. In particular, the method is inexpensive as it 
\begin{enumerate*}[label=(\itshape\alph*\upshape)]
	\item avoids the generation and examination of unnecessary (non-discriminating) or duplicate query candidates, 
	\item \emph{actually} computes only the \emph{single} best query by its ability to estimate a query's quality without computing it, and 
	\item guarantees soundness and completeness wrt.\ an exponential query search space independently of the properties and output of a reasoner.
\end{enumerate*}
Modern SQD methods like~\cite{dekleer1987} and its derivatives \cite{DBLP:journals/jair/FeldmanPG10a,Shchekotykhin2012,Rodler2013} do not meet all properties (a) -- (c) and extensively call a reasoner for (precomputed) inferences while computing a query. 
%
Moreover, by the generality of our query notion, our method explores a more complex search space than~\cite{dekleer1987,dekleer1993}, thereby guaranteeing property (5) above.

\begin{table}[t]
	\renewcommand\arraystretch{1.15}
	\scriptsize
	\centering
	\begin{tabular}{lll} 
		\hline
		$\sd$ & \multicolumn{2}{l}{$\setof{\lnot\ab(c_i) \rightarrow beh(c_i) \mid c_i \in \comps}$} \\
		\hline
		$\comps$ & \multicolumn{2}{l}{$\setof{c_1,c_2,c_3,c_4,c_5}$} \\
		\hline
		\multirow{3}{*}{\parbox{2cm}{normal behavior $beh(c_i)$ of components $c_i \in \comps$}}  & $beh(c_1): A \rightarrow B \land L$ & $beh(c_2): A \to F$  \\
		& $beh(c_3): B \lor F \to H$  		& $beh(c_4): L \to H$ 	\\ 
		&	$beh(c_5):\lnot H \to G \land \lnot A$ & 	\\
		\hline
		$\Tn$  &  \multicolumn{1}{l|}{$\tn_1 : \setof{A \to H}$} & \multicolumn{1}{|l}{$\obs, \Tp = \emptyset$} \\
		\hline
	\end{tabular}
\vspace{-4pt}
	\caption{\footnotesize Running Example DPI $\mathsf{Ex}$} 
	\label{tab:example_dpi_1}
\end{table}

	\section{Preliminaries}\label{sec:basics}
	\paragraph{Model-Based Diagnosis (MBD).}
	In this section we recap on important MBD concepts and draw on definitions of \cite{Reiter87} to characterize a system and diagnoses. \\
	\textbf{Notation (*):} Let $X$ be a collection of sets, then $U_X$ and $I_X$ denote the union and intersection of all elements of $X$, resp.
	$K \models S$ for a set $S$ is a shorthand for $K \models s$ for all $s \in S$.\qed
	
	A \emph{system} 
	consists of a set of components \comps and a system description \sd 
	where $\{\lnot\ab(c) \rightarrow beh(c) \mid c \in \comps\} \subseteq \sd$. The first-order sentence $beh(c)$ describes the normal behavior of $c$ and $\ab$ is a distinguished abnormality predicate.
	Any behavior different from $beh(c)$ implies that component $c$ is at fault, i.e.\ $\ab(c)$ holds.\footnote{We make the \emph{stationary health assumption} \cite{DBLP:journals/jair/FeldmanPG10a}: behavior of each $c \in \comps$ is constant during diagnosis.} Note, $\sd \cup \{\lnot\ab(c) \mid c\in\comps\}$ is required to be \emph{consistent}.
	
	From the viewpoint of system diagnosis, evidence about the system behavior in
	terms of observations \obs, positive ($\Tp$) and negative ($\Tn$) measurements \cite{Reiter87,dekleer1987,DBLP:journals/ai/FelfernigFJS04} is of interest. 
	
	
	\begin{definition}[DPI]\label{def:diagnosis_problem}
		Let \comps be a finite set of constants and $\sd$, $\obs$, all $\tp \in \Tp$, all $\tn \in \Tn$ be finite sets of consistent first-order sentences.
		Then $(\sd,\comps$, $\obs, \Tp,\Tn)$ is a \emph{diagnosis problem instance (DPI)}.
	\end{definition}

\begin{definition}\label{def:sdaa}
	Let $(\sd,\comps, \obs, \Tp,\Tn)$ be a DPI. Then $\sdaa{\dg} := \sd  \cup \obs \cup U_\Tp \cup \{\ab(c) \mid c\in\dg\} \cup
	\{\lnot\ab(c) \mid c\in\comps\setminus\dg\}$ denotes the behavior description of a system $(\sd,\comps)$ given observations $\obs$, union of positive measurements $U_\Tp$ as well as that all components $\dg \subseteq \comps$ are faulty and all components in $\comps \setminus \dg$ are healthy.
\end{definition} 
The solutions of a DPI, i.e.\ the hypotheses that explain a given (faulty) system behavior, are called diagnoses:
\begin{definition}[Diagnosis]\label{def:diagnosis}
	$\dg \subseteq \comps$ is a \emph{diagnosis} for the DPI $(\sd,\comps,\obs, \Tp,\Tn)$ iff $\dg$ is $\subseteq$-minimal such that
	\begin{itemize}
		\item $\sdaa{\dg}$ is consistent ($\dg$ explains $\obs$ and $\Tp$), and
		\item $\forall n \in \Tn: \sdaa{\dg} \not\models n$ ($\dg$ explains $\Tn$).
	\end{itemize}
	We denote the set of all diagnoses for a DPI $X$ by $\allD_{X}$.
\end{definition}	
%
	A diagnosis for a DPI exists iff $\sdaa{\comps} \not\models n$ for all $n \in N$ \cite[Prop.~1]{friedrich2005gdm}.
\noindent\textbf{Example:} Consider DPI $\mathsf{Ex}$ (Tab.~\ref{tab:example_dpi_1}). Using e.g.\ 
$\textsc{HS-Tree}$ \cite{Reiter87} we get (denoting components $c_i$ by $i$) the set of all diagnoses $\allD_{\mathsf{Ex}} = \{\dg_1,\dg_2,\dg_3\} = \{\{1,2,5\}, \{1,3,5\}$, $\{3,4,5\}\}$. E.g.\ $\dg_2 \in \allD_{\mathsf{Ex}}$ due to Def.~\ref{def:diagnosis} and as $\sdaa{\dg_2} = [\sd \cup \{\ab(c_1),\ab(c_3),\ab(c_5)\} \cup
\{\lnot\ab(c_2),\lnot\ab(c_4)\}] \cup \obs \cup U_\Tp = [\{beh(c_2), beh(c_4)\}] \cup \emptyset \cup \emptyset = \{A\to F, L \to H\} \not\models$ $\{A\to H\} = n_1 \in \Tn$ and is consistent.\qed
	
	\noindent\textbf{Sequential Diagnosis (SQD).}
	Given multiple diagnoses for a DPI, SQD techniques 
	extend the sets $P$ and $N$ by asking a user or an oracle (e.g.\ an automated system) to perform additional measurements in order to rule out irrelevant diagnoses.
	In line with the works of \cite{settles2012,Shchekotykhin2012,Rodler2015phd} we call a proposed measurement \emph{query} and define it very generally as a set of first-order sentences (this subsumes the notion of measurement e.g.\ in \cite{dekleer1987,Reiter87}). 
	The task of the oracle is to assess the correctness of the sentences in the query, thereby providing the required measurements. A query $Q$ is  
	$\true$ ($t$) if all sentences in $Q$ are correct and $\false$ ($f$) if at least one sentence in $Q$ is incorrect.
	
	Usually only a small computationally feasible set of \emph{leading diagnoses} $\LD$ (e.g.\ minimum cardinality \cite{DBLP:journals/jair/FeldmanPG10a} or most probable \cite{DBLP:conf/aaai/Kleer91} ones) are exploited for measurement selection~\cite{DBLP:conf/ijcai/KleerW89}. 
	
	Any sets of diagnoses and first-order sentences satisfy:
	\begin{property}\label{pty:partition}
	Let $X$ be a set of first-order sentences and $\LD \subseteq \allD_{\mathsf{DPI}}$ for $\mathsf{DPI} = (\sd,\comps,\obs,P,N)$. Then $X$ induces a partition $\Pt_\LD(X) := \tuple{\dx{}(X),\dnx{}(X),\dz{}(X)}$ on $\LD$ where $\dx{}(X) := \{\dg \in \LD \mid \sdaa{\dg} \models X\}$, $\dnx{}(X) := \{\dg \in \LD \mid \exists s \in N \cup \setof{\bot}: \sdaa{\dg} \cup X \models s\}$ and $\dz{}(X) = \LD \setminus (\dx{}(X) \cup \dnx{}(X))$.
	\end{property}
	From a query, we postulate two properties, it must for any outcome \textbf{(1)}~invalidate at least one diagnosis (\emph{search space restriction}) and \textbf{(2)}~preserve the validity of at least one diagnosis (\emph{solution preservation}). 
	In fact, the sets $\dx{}(X)$ and $\dnx{}(X)$ are the key in deciding whether a set of sentences $X$ is a query or not.
	Based on Property~\ref{pty:partition}, we define:
	\begin{definition}[Query, q-Partition]~\label{def:query}
	Let $\mathsf{DPI} = (\sd, \comps$, $\obs, P, N)$, $\LD \subseteq \allD_{\mathsf{DPI}}$ and $Q$ be a set of {first-order} sentences with $\Pt_\LD(Q) = \tuple{\dx{}(Q),\dnx{}(Q),\dz{}(Q)}$. Then $Q$ is a \emph{query} for $\LD$ 
	iff $Q\neq \emptyset$, $\dx{}(Q) \neq \emptyset$ and $\dnx{}(Q) \neq \emptyset$. The set of all queries for $\LD$ is denoted by $\mQ_{\LD}$.
	
	$\Pt_\LD(Q)$ is called \emph{the q-partition (QP)} of $Q$ iff $Q$ is a query. Inversely, $Q$ is called \emph{a query with (or: for) the QP} $\Pt_\LD(Q)$.
	
	Given a QP $\Pt$, we sometimes denote its three entries in turn $\dx{}(\Pt)$, $\dnx{}(\Pt)$ and $\dz{}(\Pt)$. 
	\end{definition}
	
	$\dx{}(Q)$ and $\dnx{}(Q)$ denote those diagnoses in $\LD$ consistent only with $Q$'s positive and negative outcome, respectively, and $\dz{}(Q)$ those consistent with both outcomes.
	Since $Q \in \mQ_{\LD}$ implies that both $\dx{}(Q)$ and $\dnx{}(Q)$ are non-empty, clearly $Q$'s outcomes both dismiss and preserve at least one diagnosis. 
	Note, in many cases a query also invalidates some (unknown) non-leading diagnoses $\allD_{\mathsf{DPI}} \setminus \LD$.
	
	We point out that the size of the set $\dz{}(Q)$ (the diagnoses that cannot be eliminated given any outcome) should be minimal, i.e.\ zero at best, for optimal diagnoses discrimination. The algorithm presented hereafter guarantees the computation of only $Q$'s with $\dz{}(Q) = \emptyset$. For example, the methods of   \cite{dekleer1987,Shchekotykhin2012,Rodler2013} cannot ensure this important property.
	
	\noindent\textbf{Example (cont'd):} Let $\LD = \allD_\mathsf{Ex} = \setof{\dg_1,\dg_2,\dg_3}$. Then, $Q = \setof{F \to H}$ is a query in $\mQ_\LD$. To verify this, let us consider its QP $\Pt_\LD(Q) = 
	\tuple{\setof{\dg_1},\setof{\dg_2,\dg_3},\emptyset}$. Since both $\dx{}(Q)$ and $\dnx{}(Q)$ are non-empty, $Q$ is in $\mQ_\LD$. $\dg_1 = \setof{1,2,5} \in \dx{}(Q)$ holds as $\sdaa{\dg_1} \models \setof{beh(c_3),beh(c_4)} = \setof{B\lor F \to H, L\to H}$ which in turn entails $Q$.
	On the other hand, e.g.\ $\dg_2 = \setof{1,3,5} \in \dnx{}(Q)$ since $\sdaa{\dg_2} \cup Q \models \setof{A\to F, L \to H, F \to H} \models \setof{A \to H} = \tn_1 \in N$.
	Hence, the outcome $Q = t$ implies that diagnoses in $\dnx{}(Q) = \setof{\dg_2,\dg_3}$ are invalidated, whereas $Q = f$ causes the dismissal of $\dx{}(Q) = \setof{\dg_1}$.\qed
		
	\noindent\textbf{Applicability and Diagnostic Accuracy.}
	For any non-singleton set of leading diagnoses, a discriminating query exists \cite[Sec.~7.6]{Rodler2015phd}: 
	\begin{property}
		$\forall \mathsf{DPI}: \LD \subseteq \allD_{\mathsf{DPI}}, |\LD| \geq 2 \implies \mQ_{\LD} \neq \emptyset$.
	\end{property}
	This has two implications: First, we need only precompute two diagnoses to generate a query and proceed with SQD. Despite its NP-completeness \cite{Bylander1991}, the generation of two (or more) 
	diagnoses is practical in many real-world settings \cite{DBLP:conf/aaai/Kleer91,Shchekotykhin2014}, making query-based SQD commonly applicable. Second, the query-based approach guarantees perfect diagnostic accuracy, i.e.\ the unambiguous identification of the actual diagnosis. 
	
	\section{Query Optimization for Sequential MBD}\label{sec:theory}
	\noindent\textbf{Measurement Selection.}
	As argued, the (q-)partition $\Pt_\LD(Q)$ enables both the \emph{verification} whether a candidate $Q$ is indeed a query and an \emph{estimation of the impact} $Q$'s outcomes have in terms of diagnoses invalidation. And, given (component) fault probabilities, it enables to \emph{gauge the probability of observing a positive or negative query outcome} \cite{dekleer1987}. 
	%
	%
	%
	%
	Active learning \emph{query selection measures (QSMs)} $m: Q \mapsto m(Q) \in \mathbb{R}$ \cite{settles2012} use exactly these query properties characterized by the QP to assess how favorable a query is. They aim at selecting queries such that the expected number of queries until obtaining a deterministic diagnostic result is minimized, i.e.\ $\sum_{\dg \subseteq \comps} p(\dg) q_{\#}(\dg) \rightarrow \min$
	where $p(\dg)$ is the (a-priori) probability that $\setof{\ab(c) \mid c \in \dg} \cup \setof{\lnot \ab(c) \mid c \in \comps\setminus\dg}$ is the actual system state wrt.\ component functionality and $q_{\#}(\dg)$ is the number of queries required, given the initial DPI, to derive that $\dg$ must be the actual diagnosis. 
	Solving this problem
	is known to be NP-complete as it amounts to optimal binary decision tree construction \cite{hyafil1976}. Hence we restrict our algorithm to the usage of QSMs that make a locally optimal query selection through a \emph{one-step lookahead}. This has been shown to be optimal in many cases and nearly optimal in most cases \cite{dekleer1992}.  
	Several different QSMs $m$ such as split-in-half, entropy, or risk-optimization have been proposed, well studied and compared against each other~\cite{dekleer1987,Shchekotykhin2012,Rodler2013}. E.g.\ using entropy as QSM, $m$ would be exactly the scoring function $\$()$ derived in \cite{dekleer1987}. Note, we assume w.l.o.g.\ that the optimal query wrt.\ any $m$ is the one with \emph{minimal} $m(Q)$.
	
Besides minimizing the number of queries in a diagnostic session, a further goal can be the minimization of the query cost (e.g.\ time, manpower). 
	To this end, one can specify a \emph{query cost measure (QCM)} $c: Q \mapsto c(Q) \in \mathbb{R}^+$. Examples of QCMs are $c_{\Sigma}(Q) := \sum_{i=1}^{k} c_i$ (prefer query with minimal overall cost, e.g.\ when $c_i$ represents time) or $c_{\max}(Q) := \max_{i \in \setof{1,\dots,k}} c_i$\label{etc:c_max} (prefer query with minimal maximal cost of a single measurement, e.g.\ when $c_i$ represents human cognitive load) where $Q = \setof{q_1,\dots,q_k}$ and $c_i$ is the cost of evaluating the truth of the first-order sentence $q_i$. The QCM $c_{|\cdot|}(Q) = |Q|$ is a special case of $c_{\Sigma}(Q)$ where $c_i = c_j$ for all $i,j$ is assumed.
	Now, the problem addressed in this work is:
\begin{problem}\label{prob:query_optimization}
	\textbf{Given:} $\mathsf{DPI} := (\sd,\comps,\obs,P,N)$, $\LD \subseteq \allD_{\mathsf{DPI}}$ with $|\LD|\geq 2$, QSM $m$, 
	QCM $c$, 
	query search space $\mathbf{S}\subseteq\mQ_{\LD}$. \textbf{Find:} A query $Q^*$ satisfying 
	$Q^* = \argmin_{Q \in \mathbf{OptQ}(m,\mathbf{S})} c(Q)$ where $\mathbf{OptQ}(m,\mathbf{S}) := \setof{Q' \mid Q' = \arg\min_{Q\in\mathbf{S}} m(Q)}$, i.e.\ $Q^*$ has 
	minimal cost wrt.\ $c$ among all queries in $\mathbf{S}$ that are 
	optimal wrt.\ $m$. 
\end{problem}
Note there can be multiple equally good queries $Q^* \in \mQ_{\LD}$.

	\noindent\textbf{The Algorithm}
	we propose to solve Problem~\ref{prob:query_optimization} is given by Alg.~\ref{algo:query_comp}. The described query computation procedure can be divided into three phases: P1 
	(line~\ref{algoline:query_comp:optimizeQP}), P2 
	(line~\ref{algoline:query_comp:optimizeQforQP}) and (optionally) P3 
	(lines~\ref{algoline:query_comp:enrichQ}-\ref{algoline:query_comp:optiminimizeQ}). We next give the intuition and explanation of these phases. 
	
	\begin{algorithm}[t]
		\scriptsize
		\caption{Optimized Query Computation}\label{algo:query_comp}
		\begin{algorithmic}[1]
			\Require $\mathsf{DPI}:=(\sd,\comps,\obs,P,N)$, $\LD \subseteq \allD_{\mathsf{DPI}}$, $|\LD|\geq 2$, QSM $m$, QCM $c$, component fault probabilities $\mathit{FP} = \setof{p_i \mid p_i = p(c_i), c_i \in \comps}$,
			threshold $t_m$ 
			(i.e.\ $|m(Q) - m_{opt}| \leq t_m \Rightarrow Q$ regarded optimal; $m_{opt} :=$ optimal value of $m$),
			sound and complete inference engine $\mathit{Inf}$, set $\mathit{ET}$ of entailment types 
			\Ensure an optimized query $Q^* \in \mQ_{\mD}$ wrt.\ $m$, $t_m$ and $c$	(cf.\ Theorems~\ref{theorem:P1+P2_solve_query_optimization_problem} and \ref{theorem:P3_solves_problem_1})		
			\State $\Pt \gets \Call{optimizeQPartition}{\mD,\mathit{FP},m,t_m}$ \label{algoline:query_comp:optimizeQP}    \Comment{P1}
			\State $Q^* \gets \Call{optimizeQueryForQPartition}{\Pt,\mathit{FP},c}$ \label{algoline:query_comp:optimizeQforQP}   \Comment{P2}
			\If{$\mathsf{enhance} = \true$}
			\State $Q' \gets \Call{expandQuery}{\mathsf{DPI}, \Pt, \mathit{Inf}, ET}$    \Comment{(optional) P3} \label{algoline:query_comp:enrichQ}
			\State $Q^* \gets \Call{optiMinimizeQuery}{Q',\Pt,\mathsf{DPI},\mathit{FP}, \mathit{Inf}}$	 \Comment{(optional) P3}	\label{algoline:query_comp:optiminimizeQ}
			\EndIf
			\State \Return $Q^*$
		\end{algorithmic}
		\normalsize
	\end{algorithm}
	
	\noindent\textbf{Phase P1.} At this stage, we optimize the given QSM $m$ -- for now without regard to the QCM $c$, which is optimized later in P2. This \emph{decoupling of optimization steps} is possible since the QSM value $m(Q)$ of a query $Q$ is only affected by the (unique) QP of $Q$ and not by $Q$ itself. On the contrary, the QCM value $c(Q)$ is a function of $Q$ only and not of $Q$'s QP.
	Therefore, the search performed in P1 will consider only QPs.
	
	To verify whether a given $3$-partition of $\LD$ is a QP, however, we need a query $Q$ for this QP which lets us determine whether $\dx{}(Q) \neq \emptyset$ and $\dnx{}(Q) \neq \emptyset$ (cf.\ Def.~\ref{def:query}). But:
	\begin{property}\label{pty:query_vs_QP}
		For one query there is exactly one QP (immediate from Property~\ref{pty:partition}). For one QP there might be an exponential number of queries (cf.\ Propos.~\ref{prop:number_of_CQPs} later).
	\end{property}	
	Therefore, we use the notion of a \emph{canonical query (CQ)}, which is one well-defined query representative for a QP. From a CQ, we postulate \emph{easiness of computation} and \emph{exclusion of suboptimal QPs} with $\dz{} \neq \emptyset$ (cf.\ Sec.~\ref{sec:basics}). The key to realizing these postulations is:
	\begin{definition}\label{def:BEH}
	$X \subseteq \comps$, $\beh{X} := \setof{beh(c_i)\mid c_i \in X}$. 
	\end{definition}
	The following property is immediate from Def.~\ref{def:sdaa}:
	\begin{property}\label{pty:sdaa_models_BEH}
	$X \subseteq \comps \implies \sdaa{X} \models \beh{\comps \setminus X}$
	\end{property}
	From Property~\ref{pty:partition} and Def.~\ref{def:query} we can directly conclude:
	\begin{property}\label{pty:query_is_subset_of_common_entailments}
		A query $Q \in \mQ_{\LD}$ is a subset of the common entailments of all KBs in the set $\setof{\sdaa{\dg} \mid \dg \in \dx{}(Q)}$.
	\end{property}
	Using Properties~\ref{pty:sdaa_models_BEH} and \ref{pty:query_is_subset_of_common_entailments}, the idea is now to restrict the space of entailments of the $\sdaa{\cdot}$ KBs to the behavioral descriptions $beh(\cdot)$ of the system components. That is, each CQ should be some query $Q \subseteq \beh{\comps}$. This assumption along with Def.~\ref{def:query} and the $\subseteq$-minimality of diagnoses yields:
	\begin{proposition}\label{prop:beh_query_must_mustnot_neednot}
		Any query $Q \subseteq \beh{\comps}$ in $\mQ_\LD$ \emph{must} include some formulas in $\beh{U_\LD}$, \emph{need not} include any formulas in $\beh{\comps \setminus U_\LD}$, and \emph{must not} include any formulas in $\beh{I_\LD}$. (Please refer to (*) in Sec.~\ref{sec:basics} for notation.)	
		
		Moreover, the deletion of any sentences in $\beh{\comps \setminus U_\LD}$ from $Q$ does not alter the QP $\Pt_\LD(Q)$.
	\end{proposition} 
	Hence, we define:
	\begin{definition}\label{def:discax}
		$\Disc_\LD := \beh{U_\LD} \setminus \beh{I_\LD} = \beh{U_\LD \setminus I_\LD}$ the \emph{discrimination sentences wrt.\ $\LD$} (i.e.\ those essential for discrimination between diagnoses in $\LD$).
	\end{definition}
	CQs can now be characterized as follows:
	\begin{definition}[CQ]\label{def:CQ}
		Let $\emptyset\subset\dx{}\subset\LD$. Then $Q_{\mathsf{can}}(\dx{}) := \beh{\comps \setminus U_{\dx{}}}$  $\cap \Disc_\LD$ is \emph{the canonical query (CQ)} wrt.\ seed $\dx{}$ if $Q_{\mathsf{can}}(\dx{}) \neq \emptyset$. Else, $Q_{\mathsf{can}}(\dx{})$ is undefined.
	\end{definition}
	Note, $\beh{\comps \setminus U_{\dx{}}}$ are exactly the common $beh(\cdot)$ entailments of $\setof{\sdaa{\dg} \mid \dg \in \dx{}}$ (cf.\ Property~\ref{pty:query_is_subset_of_common_entailments}). 
	The CQ extracts 
	$\Disc_\LD$
	from these entailments, thereby removing all elements that do not affect the QP (cf.\ Propos.~\ref{prop:beh_query_must_mustnot_neednot}). By Defs.~\ref{def:query} and \ref{def:CQ} and the $\subseteq$-minimality of diagnoses, we get:
 	\begin{proposition}\label{prop:CQ_is_query}
	If $Q$ is a CQ, then $Q$ is a query.
	\end{proposition}
	
	The QP for a CQ is called canonical q-partition:
	\begin{definition}[CQP]\label{def:CQP}
		A QP $\Pt'$ for which a CQ $Q$ exists with QP $\Pt'$, i.e.\ $\Pt(Q) = \Pt'$, is called a \emph{canonical QP (CQP)}.
	\end{definition}
	Since a CQ is a subset of $\beh{\comps}$ and diagnoses are $\subseteq$-minimal, we can derive:
	\begin{proposition}\label{prop:CQPs_have_empty_dz}
	Let $\Pt$ be a CQP. Then $\dz{}(\Pt) = \emptyset$.
	\end{proposition} 
	\noindent\textbf{Discussion:}
	The restriction to CQs during P1 has some nice implications: \textbf{(1)}~CQs can be generated by cheap set operations (\emph{no inference engine calls}), \textbf{(2)}~each CQ is a query in $\mQ_\LD$ for sure (Propos.~\ref{prop:CQ_is_query}), no verification of its QP (as per Def.~\ref{def:query}) required, thence \emph{no unnecessary} (non-query) \emph{candidates generated},
	\textbf{(3)}~\emph{automatic focus on favorable queries} wrt.\ the QSM $m$ (those with empty $\dz{}$, Propos.~\ref{prop:CQPs_have_empty_dz}),  
	\textbf{(4)}~\emph{no duplicate QPs generated} as there is a one-to-one relationship between CQs and CQPs (Property~\ref{pty:query_vs_QP}, Def.~\ref{def:CQ}), 
	\textbf{(5)} the explored search space for QPs is \emph{not dependent on} the particular (entailments) output by an \emph{inference engine}.
	
	We emphasize that all these properties do \textbf{not} hold for normal (i.e.\ non-canonical) queries and QPs. The overwhelming impact of this will be demonstrated in Sec.~\ref{sec:eval}.\qed
	
	\noindent\textbf{Example (cont'd):} 
	Given $\LD$ as before, $\Disc_\LD = \beh{U_\LD \setminus I_\LD} = \beh{\setof{1,2,3,4,5}\setminus\setof{5}} = \beh{\setof{1,2,3,4}}$. Let us consider the seed $\dx{} = \setof{\dg_1} = \setof{\setof{1,2,5}}$. Then the CQ $Q_1 := Q_{\mathsf{can}}(\dx{}) = (\beh{\setof{1,2,3,4,5} \setminus \setof{1,2,5}}) \cap \beh{\setof{1,2,3,4}} = \beh{\setof{3,4}}$. The associated CQP is $\Pt_1 = \tuple{\setof{\dg_1},\setof{\dg_2,\dg_3},\emptyset}$. Note, $\dg \in \dx{}(\Pt_1)$ ($\dg \in \dnx{}(\Pt_1)$) for a $\dg \in \LD$ iff $\beh{\comps \setminus \dg} \supseteq (\not\supseteq ) Q_1$. E.g.\ $\dg_3\in\dnx{}(\Pt_1)$ since $\beh{\comps \setminus \dg_3} = \beh{\setof{1,2}} \not\supseteq \beh{\setof{3,4}} = Q_1$. That is, using CQs and CQPs, reasoning is traded for set operations and comparisons.
	
	The seed $\dx{} = \setof{\dg_1,\dg_3}$ yields $Q_2 := Q_{\mathsf{can}}(\dx{})= (\beh{\setof{1,\dots,5} \setminus \setof{1,\dots,5}}) \cap \beh{\setof{1,\dots,4}} = \emptyset$, i.e.\ there is no CQ wrt.\ seed $\dx{}$ and the partition $\tuple{\setof{\dg_1,\dg_3},\setof{\dg_2},\emptyset}$ with the seed $\dx{}$ as first entry is no CQP (and also no QP).
	\qed

	Now, having at hand the notion of a CQP, we describe the (heuristic) depth-first, local best-first (i.e.\ chooses only among best direct successors at each step) backtracking CQP search procedure performed in P1.
	
	A (heuristic) search problem \cite{russellnorvig2010} is defined by the \emph{initial state}, a \emph{successor function} enumerating all direct neighbor states of a state, the \emph{step costs} from a state to a successor state, the \emph{goal test} to determine if a given state is a goal state or not, (and some \emph{heuristics} to estimate the remaining effort towards a goal state). 
	
	We define the initial state 
	$\langle\dx{},\dnx{},\dz{}\rangle$ 
	as $\langle\emptyset,\LD,\emptyset\rangle$.
	The idea is to transfer diagnoses step-by-step from $\dnx{}$ to $\dx{}$ to construct all CQPs \emph{systematically}. The step costs are irrelevant, only the found QP as such counts.
	Heuristics derived from the QSM $m$ (cf.\ e.g.\ \cite{Shchekotykhin2012}) can be (optionally) integrated into the search to enable faster convergence to the optimum.
	A QP is 
	a goal if it optimizes $m$ up to the given threshold $t_m$ (cf.\ \cite{dekleer1987}, see Alg.~\ref{algo:query_comp}). In order to characterize a suitable successor function, we define a direct neighbor of a QP as follows:
	\begin{definition}\label{def:minimal_transformation}
		Let $\Pt_i := \langle \dx{i},\dnx{i},\emptyset\rangle$,
		$\Pt_j := \langle \dx{j},\dnx{j},\emptyset\rangle$ be partitions of $\LD$. 
		Then, $\Pt_i \mapsto \Pt_j$ is a \emph{minimal $\dx{}$-transformation} from $\Pt_i$ to $\Pt_j$ iff $\Pt_j$ is a CQP, $\dx{i} \subset \dx{j}$ and there is no CQP $\langle \dx{k},\dnx{k},\emptyset\rangle$ with $\dx{i} \subset \dx{k} \subset \dx{j}$. 
		
		A CQP $\Pt'$ is called a \emph{successor of a partition $\Pt$} iff $\Pt'$ results from $\Pt$ by a minimal $\dx{}$-transformation.
	\end{definition}
	For the initial state successors we get~\cite[p.~98]{Rodler2015phd}:
	\begin{proposition}\label{prop:succ_of_initial_state}
		The CQPs $\langle\setof{\dg},\LD\setminus\setof{\dg},\emptyset\rangle$ for $\dg \in \LD$ are exactly all successors of $\tuple{\emptyset,\LD,\emptyset}$.
	\end{proposition}
	To specify the successors of an intermediate CQP $\Pt_k$ in the search, we draw on diagnoses' \emph{traits}:
	\begin{definition}
		Let $\Pt_k = \langle \dx{k}, \dnx{k}, \emptyset\rangle$ be a CQP and $\dg_i \in \dnx{k}$. Then the \emph{trait} $\dg_i^{(k)}$ of $\dg_i$ is defined as $\beh{\dg_i \setminus U_{\dx{k}}}$. 
	\end{definition}
	The relation $\sim_k$ associating two diagnoses in $\dnx{k}$ iff their trait is equal is obviously an equivalence relation. Now, Defs.~\ref{def:CQ}, \ref{def:CQP} and \ref{def:minimal_transformation} let us derive: 
	\begin{proposition}\label{prop:succ_of_intermediate_state}
	Let $\mathit{EC} := \setof{E_1,\dots,E_s}$ be the set of all equivalence classes wrt.\ $\sim_k$. $\Pt_k$ has successors iff $s \geq 2$. In this case, all successors are given by $\tuple{\dx{k} \cup E,\dnx{k}\setminus E,\emptyset}$ where $E \in \mathit{EC}$ and $E$ has a $\subseteq$-minimal trait among all classes $E' \in \mathit{EC}$. 
	\end{proposition}
	 By Def.~\ref{def:minimal_transformation} which demands both \emph{minimal changes} between state and successor state and \emph{the latter to be a CQP}, we have:
	 \begin{theorem}\label{theorem:CQP_search_sound+complete}
	 	Usage of the successor function as given in Propos.~\ref{prop:succ_of_initial_state} (for initial state) and Propos.~\ref{prop:succ_of_intermediate_state} (for intermediate states) makes the search for CQPs sound and complete.
	 \end{theorem}
	Since it can be proven that $\Pt = \tuple{\dx{},\dnx{},\emptyset}$ is a CQP iff $U_{\dx{}} \subset U_{\mD}$ and as there are at least $|\LD|$ CQPs (Propos.~\ref{prop:succ_of_initial_state}): 
	\begin{proposition}\label{prop:number_of_CQPs}
		Let $\mathbf{CQP}_\LD$ denote the set of CQPs for diagnoses $\LD$ with $|\LD|\geq 2$.
		Then $|\mathbf{CQP}_\LD| = |\{U_{\dx{}}\,|\,\emptyset\subset\dx{}\subset\mD, U_{\dx{}} \neq U_\mD\}| \geq |\mD|$.
	\end{proposition}
	%
Whether QPs $\langle\dx{},\dnx{},\emptyset\rangle$ exist which are no CQPs is not yet clarified, but both
theoretical and empirical evidence indicate the negative.
E.g., an analysis of $\approx 900\,000$ QPs we ran for different diagnoses $\LD$ and DPIs showed that \emph{all QPs were indeed CQPs}. 
And, in all evaluated cases (see Sec.~\ref{sec:eval}) optimal CQPs 
wrt.\ all QSMs $m$ given in diagnosis literature \cite{dekleer1987,Shchekotykhin2012,Rodler2013} were found. Hence:
%
	\begin{conjecture}\label{conj:CQPs=QPs}
	Let $\mathbf{(C)QP}_\LD$ 
	denote the sets of (C)QPs 
	(all with $\dz{}=\emptyset$) for diagnoses $\LD$. Then $\mathbf{CQP}_\LD = \mathbf{QP}_\LD$.
	\end{conjecture}
	
	\vspace{2pt}
	
	\noindent\textbf{Example (cont'd):} Reconsider the CQP $\Pt_1 = \langle\setof{\dg_1}, \{\dg_2$, $\dg_3\}, \emptyset\rangle$.
	The traits are $\dg_2^{(1)} = \beh{\{1,3,5\} \setminus \{1,2,5\}} =$ $ \beh{\setof{3}}$ and $\dg_3^{(1)} = \beh{\setof{3,4}}$, representing two equivalence classes wrt.\ $\sim_1$. There is only one class with $\subseteq$-minimal trait, i.e.\ $\setof{\dg_2}$. Hence, there is just a single successor CQP $\Pt_2 = \langle\{\dg_1,\dg_2\}$, $\{\dg_3\},\emptyset\rangle$ of $\Pt_1$. Recall, we argued that $\langle\{\dg_1,\dg_3\}$, $\{\dg_2\},\emptyset\rangle$ is indeed no CQP. 
	By Propos.~\ref{prop:number_of_CQPs}, there are $|\{\setof{1,2,5}, \setof{1,3,5}, \setof{3,4,5}, \setof{1,2,3,5}$, $\setof{1,3,4,5}\}| = 5$ different CQPs wrt.\ $\LD$. 
	Note, Conject.~\ref{conj:CQPs=QPs} is true here, 
	i.e.\ the $\mathbf{CQP}_\LD$ search is complete wrt.\ $\mathbf{QP}_\LD$.
	\qed

%
%
%
%
%
%
%
%
%

	\noindent\textbf{Phase P2.}
	Phase P1 returns an optimal (C)QP $\Pt_k$ wrt.\ the QSM $m$. Property~\ref{pty:query_vs_QP} indicates that there might be still a large search space for an optimal query wrt.\ the QCM $c$ for this QP. The task in P2 is to find such query efficiently.
	
	From $\Pt_k$, we can obtain the associated CQ $Q_k$ (as per Def.~\ref{def:CQ}). However, usually a least requirement of any QCM $c$ is i.a.\ the $\subseteq$-minimality of a query to avoid unnecessary measurements. To this end, let $\mathsf{Tr}(\Pt_k)$ denote the set of all $\subseteq$-minimal traits wrt.\ $\sim_k$. Given a collection of sets $X = \setof{x_1,\dots,x_n}$, a set $H \subseteq U_X$ is a \emph{hitting set (HS)} of $X$ iff $H \cap x_i \neq \emptyset$ for all $x_i \in X$. Then: 
	\begin{proposition}\label{prop:min_query_is_min_HS}
		$Q \subseteq \Disc_\mD$ is a $\subseteq$-minimal query with QP $\Pt_k$ iff $Q = H$ for some $\subseteq$-minimal HS $H$ of $\mathsf{Tr}(\Pt_k)$. 
	\end{proposition}

	Hence, all $\subseteq$-minimal reductions of CQ $Q_k$ under preservation of the (already fixed and optimal) QP $\Pt_k$ can be computed e.g.\ using the classical $\textsc{HS-Tree}$ \cite{Reiter87}. However, there is a crucial difference to standard application scenarios of $\textsc{HS-Tree}$, namely the fact that all sets to label the tree nodes (i.e.\ the $\subseteq$-minimal traits) are readily available (without further computations). Consequently, the construction of the tree runs swiftly, as our evaluation will confirm. Note also, in principle we only require a \emph{single} minimal hitting set, i.e.\ query. Moreover, $\textsc{HS-Tree}$ can be used as uniform-cost (UC) search (cf.\ e.g.\ \cite[Chap.~4]{Rodler2015phd}), incorporating
	the QCM $c$ to find queries in best-first order wrt.\ $c$. 
	In fact, all QCMs (i.e.\ $c_{\Sigma}$, $c_{\max}$, $c_{|\cdot|}$) discussed
	above can be optimized using UC $\textsc{HS-Tree}$. In case some QCM $c$ is not suitable for UC search, a brute force $\textsc{HS-Tree}$ search over all $\subseteq$-minimal queries will be practical as well (no expensive operations involved). Hence, P1 and P2 provide a solution to Problem~\ref{prob:query_optimization} \emph{without a single inference engine call}. 
	\begin{theorem}\label{theorem:P1+P2_solve_query_optimization_problem}
		P1 and P2 compute a solution $Q^*$ 
		to Problem~\ref{prob:query_optimization} where $\mathbf{S} := \setof{\beh{X} \mid X \subseteq \comps}$.
	\end{theorem}

	
	\vspace{2pt}
	
	\noindent\textbf{Example (cont'd):} Recall the CQP $\Pt_1$ and let the QCM be $c := c_{|\cdot|}$. 
	Then $\mathsf{Tr}(\Pt_1) = \setof{\beh{\setof{3}}}$, i.e.\ by Propos.~\ref{prop:min_query_is_min_HS} there is a single $c$-optimal query $\beh{\setof{3}}$ 
	for $\Pt_1$, a proper subset of the CQ $\beh{\setof{3,4}}$ for $\Pt_1$. 
	Considering the CQP $\Pt_3 := \tuple{\setof{\dg_2},\setof{\dg_1,\dg_3},\emptyset}$, $\mathsf{Tr}(\Pt_3) = \setof{\beh{\setof{2}},\beh{\setof{4}}}$ and thus we have (Propos.~\ref{prop:min_query_is_min_HS}) a single $c$-optimal query $\beh{\setof{2,4}}$ 
	which happens to be equal to the CQ for $\Pt_3$. \qed
	
	\noindent\textbf{Phase P3.}
	The query $Q^*$ optimized along two dimensions (\# of queries and cost per query) output by P2 can be directly proposed as next measurement. A $\beh{\cdot}$ query like $Q^*$ would correspond to a direct examination of one or more system components, e.g.\ to ping servers in a distributed system \cite{DBLP:conf/ijcai/BrodieRMO03}, to test gates using a voltmeter in circuits \cite{dekleer1987} or to ask the stakeholders of a (software/configuration/KB) system whether specified code lines/constraints/sentences are correct \cite{DBLP:journals/ai/Wotawa02_1,DBLP:journals/ai/FelfernigFJS04,friedrich2005gdm}.
	%
	
	Alternatively, the already optimal CQP $\Pt_k$ returned by P1 can be regarded as intermediate solution to building a solution query to Problem~\ref{prob:query_optimization} with full search space $\mathbf{S} = \mQ_{\LD}$. 
	To this end, first, using the CQ $Q_k$ of $\Pt_k$, a (finite) set $Q_{\mathsf{exp}}$ of first-order sentences of types $\mathit{ET}$ (e.g.\ atoms or sentences of type $A \to B$) are computed. $Q_{\mathsf{exp}}$ must meet: \textbf{(1)}~$\sdaa{X} \models Q_{\mathsf{exp}}$ where $X$ is some (superset of a) diagnosis such that $Q_k \subseteq \sdaa{X}$ (\emph{entailed by a consistent system behavior KB}), \textbf{(2)}~no $q_i \in Q_{\mathsf{exp}}$ is an entailment of $\sdaa{X} \setminus Q_k$ (logical dependence on $Q_k$, \emph{no irrelevant sentences}) and 
	\textbf{(3)}~the expansion of $Q_k$ by $Q_{\mathsf{exp}}$ does \emph{not alter the (already fixed and optimal) q-partition} $\Pt_k$, i.e.\ $\Pt_k = \Pt(Q_k \cup Q_{\mathsf{exp}})$.
	\begin{proposition}\label{prop:enrichment_function}
	Let $Ent_{\mathit{ET}}(X)$ be a \emph{monotonic} consequence operator realized by some inference engine that computes a finite set of entailments of types $\mathit{ET}$ of a KB $X$. Postulations \textbf{(1)} -- \textbf{(3)} are satisfied if $Q_{\mathsf{exp}} := Ent_{\mathit{ET}}(\sdaa{U_\LD} \cup Q_k) \setminus Ent_{\mathit{ET}}(\sdaa{U_\LD})$. 	
	\end{proposition}
	Finally, the expanded query $Q' := Q_k \cup Q_{\mathsf{exp}}$ can be minimized to get a $\subseteq$-minimal subset of it under preservation of the associated QP $\Pt_k$. For this purpose, one can use a variant of the polynomial divide-and-conquer method \textsc{QuickXPlain} \cite{junker04}, e.g.\ the \textsc{minQ} procedure given in \cite[p.111 ff.]{Rodler2015phd}. However, we propose to alter the input to \textsc{minQ} as follows: Assume that $Q'$ can be partitioned into a subset of \emph{cost-preferred sentences} $Q'_{\textsc{c}+}$ (e.g.\ those measurements executable automatically by available built-in sensors) and cost-dispreferred ones $Q'_{\textsc{c}-} = Q' \setminus Q'_{\textsc{c}+}$ (e.g.\ manual measurements). Let the input to \textsc{minQ} be the list $[Q'_{\textsc{c}+},asc(Q'_{\textsc{c}-})]$ (reordering of $Q'$) where $asc(Q'_{\textsc{c}-})$ means that $Q'_{\textsc{c}-}$ is sorted in ascending order by sentence cost. Then:
	\begin{proposition}\label{prop:minQ_sorted}
	$\textsc{minQ}$ with input $[Q'_{\textsc{c}+},asc(Q'_{\textsc{c}-})]$ returns a $\subseteq$-minimal query $Q^{*} \subseteq Q'$ such that $\Pt(Q^{*}) = \Pt_k$. Further, if such a query comprising only $Q'_{\textsc{c}+}$ (and no $Q'_{\textsc{c}-}$) sentences exists, then $Q^{*} \subseteq Q'_{\textsc{c}+}$. Else, $Q^{*}$ optimizes the QCM $c_{\max}$ (cf.\ page~\pageref{etc:c_max}) among all $\subseteq$-minimal subsets of $Q'$ with QP $\Pt_k$.
	\end{proposition}
	Note, phase P3, i.e.\ query expansion (Propos.~\ref{prop:enrichment_function}) together with optimized minimization (Propos.~\ref{prop:minQ_sorted}), requires only a \emph{polynomial number of inference engine calls} \cite{junker04}.
	\begin{theorem}\label{theorem:P3_solves_problem_1}
	Let Conject.~\ref{conj:CQPs=QPs} hold and the QCM be $c_{\max}$ (cf.\ page~\pageref{etc:c_max}). 
	Then P3, using the QP output by P1 and Propos.~\ref{prop:enrichment_function} and \ref{prop:minQ_sorted}, solves Problem~\ref{prob:query_optimization} with full search space $\mathbf{S} = \mQ_{\LD}$.	
	\end{theorem}
	\noindent\textbf{Example (cont'd):} Assume the QP $\Pt_1$ is returned by P1. Let the cost $c_i$ of a sentence $q_i$ be the number of literals in its clausal form. As shown before, the CQ of $\Pt_1$ is $Q_1 := \beh{\{3,4\}} = \{B \lor F \to H, L \to H\}$. Using Propos.~\ref{prop:enrichment_function} with $\mathit{ET}$ set to ``definite clauses with singleton body'', we get $Q_{\mathsf{exp}} = Ent_{\mathit{ET}}(Q_1) \setminus Ent_{\mathit{ET}}(\emptyset) = \{B \to H, F \to H, L \to H\}$. So, $Q' = \{B \to H, F \to H, L \to H, B \lor F \to H\}$. Suppose $\mathit{ET}$ defines exactly the cost-preferred sentences, i.e.\ $Q'_{c+} = Q_{\mathsf{exp}}$. Running \textsc{minQ} with input $[Q_{\mathsf{exp}},\{B \lor F \to H\}]$ yields $Q^* = \{F \to H\}$, a query that includes only cost-preferred elements (cf.\ Propos.~\ref{prop:minQ_sorted}). It is easily verified by means of Property~\ref{pty:partition} that $Q^*$ has still the QP $\Pt_1$.\qed 
	

\section{Evaluation}\label{sec:eval}
To evaluate our method, we used real-world inconsistent knowledge-based (KB) systems as \textbf{(1)}~they pose a hard challenge for query selection methods due to the implicit nature of the possible queries (must be derived by inference; not directly given such as wires in a circuit), 
\textbf{(2)}~any MBD system in the sense of \cite{Reiter87} is described by a KB,
\textbf{(3)}~the \emph{type} of the underlying system is irrelevant to our method, only its \emph{size} and \emph{(reasoning) complexity} -- for the optional phase P3 -- and the \emph{DPI structure}, e.g.\ size, \# or probability of diagnoses -- for phases P1, P2 -- are critical.
To account for this, we used systems (see Tab.~\ref{tab:experiment_data_set}, col.~1) of different size (\# of components, i.e.\ logical axioms in the KB, see Tab.~\ref{tab:experiment_data_set}, col.~2), complexity 
(see Tab.~\ref{tab:experiment_data_set}, col.~3) and DPI structure (see Tab.~\ref{tab:experiment_data_set}, col.~4).  

In our experiments, for each faulty system's DPI $\mathsf{Sys}$ in Tab.~\ref{tab:experiment_data_set} and each $n \in \setof{10,20,\dots,80}$, we randomly generated 5 different $\LD\in\allD_{\mathsf{Sys}}$ with $|\LD| = n$ using \textsc{Inv-HS-Tree} \cite{Shchekotykhin2014} with randomly shuffled input. Each $\md \in \LD$ was assigned a uniformly random probability. 

For each of these 5 $\LD$-sets, we used (a)~entropy (ENT) \cite{dekleer1987} and (b)~split-in-half (SPL) \cite{Shchekotykhin2012} 
as QSM $m$ and $c_{|\cdot|}$ (cf.\ page~\pageref{etc:c_max}) as QCM $c$, and then ran phases (i)~P1+P2 and (ii)~P3 to compute an optimized query as per Theorems~\ref{theorem:P1+P2_solve_query_optimization_problem} and \ref{theorem:P3_solves_problem_1}, respectively. We specified the optimality threshold $t_m$ as $0.01$ in (a) and $0$ in (b), cf.\ Alg.~\ref{algo:query_comp}.
The search in P1 (cf.\ Sec.~\ref{sec:theory}) used the greedy heuristic discussed in \cite[p.~11]{Shchekotykhin2012}. In P3 simple definite clauses of the form $\forall x (A(x) \to B(x))$ 
were considered cost-preferred (cf.\ last Example above).

\begin{figure}
	\centering
	\includegraphics[width=0.95\linewidth]{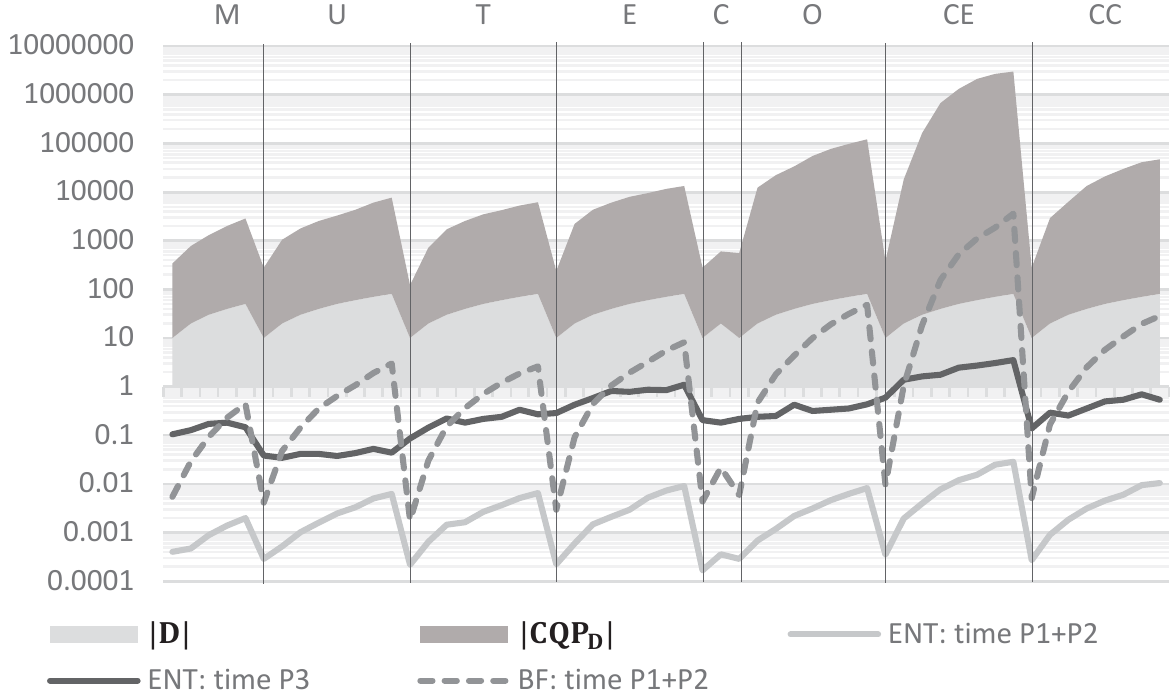}
	\vspace{-7pt}
	\caption{\footnotesize Results for systems in Tab.~\ref{tab:experiment_data_set} (x-axis): \# of leading diagnoses $|\LD|$, associated size $|\mathbf{CQP}_\LD|$ of CQP search space, and computation time~(sec) required by phases P1+P2 and P3 for QSM ENT with threshold $t_m=0.01$ and brute force (BF) search (y-axis).}
	\label{fig:eval}
\end{figure}

\noindent\textbf{Experimental Results} are shown in Fig.~\ref{fig:eval}. 
Times for SPL are omitted 
for clarity as they were quasi the same as for ENT.
The dark 
gray area shows the \# of CQPs addressed by P1, and the light gray line the time for P1+P2 using ENT. 
It is evident that P1+P2 always finished in less than $0.03$ sec outputting an optimized query wrt.\ $m$ and $c$. Note, albeit P1+P2 solve Prob.~\ref{prob:query_optimization} for a 
restricted search space $\mathbf{S}$ (cf.\ Theor.~\ref{theorem:P1+P2_solve_query_optimization_problem}), $|\mathbf{CQP}_\LD|$, a fraction of $|\mathbf{S}|$, already averaged to e.g.\ $300$ (over $|\mD| = 10$ cases) and $>530\,000$ ($|\LD|=80$). That $|\mathbf{S}|$ is sufficiently large for all sizes $|\LD|$ is also substantiated by the fact that \emph{in each single run} an optimal query wrt.\ the very small $t_m$ ($\frac{1}{10}$ of $t_m$ used in \cite{Shchekotykhin2012}) was found in $\mathbf{S}$. Also, a brute force (BF) search (dashed line) iterating over all possible CQPs is feasible in most cases -- finishing within 1~min for all runs (up to search space sizes $>120\,000$) except the $|\LD|\geq 30$ cases for system CE 
(where up to 3 million CQPs were computed). 
This extreme speed is possible due to the \emph{complete avoidance of costly reasoner calls}. The optional further query enhancement in P3 using a reasoner~\cite{sirin2007pellet} always finished within 4 sec and returned the globally optimal query wrt.\ QCM $c_{\max}$ (Theor.~\ref{theorem:P3_solves_problem_1}). 
The median output query size after P1+P2+P3 was $3.4$.
In additional scalability tests using $|\LD| = 500$ for the large enough DPIs (CC, CE, T, E) P1+P2 always ended in $< 0.6$ sec, P3 in $< 40$ sec. 
%
%
%

We also simulated P1 by a method using \emph{non-canonical} QPs, thus relying on a reasoner. For \emph{no} DPI in Tab.~\ref{tab:experiment_data_set} a result for $|\LD| > 15$ could be found in $\leq 1$~h. And, the quality of the returned QP (if any) wrt.\ $m$ was \emph{never} better than for P1.

\begin{table}[t]
\begin{threeparttable}[t]
	\renewcommand\arraystretch{1.04}
	\scriptsize
	\centering
	\begin{tabular}{llll} 
		\hline
		System 					& $|\comps|$& Complexity \tnote{a} 		& \#D/min/max \tnote{b} \\ \hline
		University (U) \tnote{c}   			& 49 		& $\mathcal{SOIN}^{(D)}$& 90/3/4      \\		
		MiniTambis (M) \tnote{c}			& 173 		& $\mathcal{ALCN}$ 		& 48/3/3	  \\
		CMT-Conftool (CC) \tnote{d}		& 458  		& $\mathcal{SIN}^{(D)}$ & 934/2/16		  \\
		Conftool-EKAW (CE) \tnote{d}		& 491 		& $\mathcal{SHIN}^{(D)}$& 953/3/10		  \\
		Transportation (T) \tnote{c}		& 1300 		& $\mathcal{ALCH}^{(D)}$& 1782/6/9	  \\
		Economy (E) \tnote{c}			& 1781 		& $\mathcal{ALCH}^{(D)}$& 864/4/8     \\
		Opengalen-no-propchains (O) \tnote{e} & 9664 	& $\mathcal{ALEHIF}^{(D)}$ & 110/2/6  \\
		Cton (C) \tnote{e}				& 33203 	& $\mathcal{SHF}$ 		& 15/1/5     \\
		\hline
\end{tabular}
\begin{tablenotes}
	\item[a] Description Logic expressivity, cf.\ \cite[p.~525 ff.]{DLHandbook}.
	\item[b] \#D, min, max denote \#, min.\ and max.\ size of all diagnoses (computable in $\leq 8$ h).
	\item[c] Sufficiently complex systems (\#D $\geq 40$) used in \cite{Shchekotykhin2012}.
	\item[d] Hardest diagnosis problems mentioned in \cite{Stuckenschmidt2008}.
	\item[e] Hardest diagnosis problems tested in \cite{Shchekotykhin2012}.
\end{tablenotes}
\vspace{-7pt}
	\caption{\footnotesize Systems used in Experiments}
	\label{tab:experiment_data_set}
\end{threeparttable}
\end{table}

	\section{Conclusion}
We present a search that addresses the optimal measurement (\emph{query}) selection problem for sequential diagnosis and is applicable to any model-based diagnosis problem conforming to \cite{dekleer1987,Reiter87}. In particular, we allow a query to be optimized along two dimensions, i.e.\ number of queries and cost per query. We show that the optimizations of these properties can be decoupled and considered in sequence. For a suitably restricted (still exponential) query search space 
(very close approximations of) 
\emph{global optima} wrt.\ given query quality measures are found \emph{without any calls to an inference engine} in \emph{negligible time} for diagnosis problems of any size and complexity 
(given the precomputation of $\geq 2$ diagnoses is feasible).
E.g.\ query search spaces of size up to 
$3$ million can be handled \emph{instantaneously} ($<0.1$ sec). For the full search space, under reasonable assumptions, the globally optimal query wrt.\ a cost-preference measure can be found within 4 sec for up to $80$ leading diagnoses.


	\newpage
	\bibliographystyle{named}
	\balance
	\bibliography{library}


\end{document}